\documentclass{article}

\usepackage{microtype}
\usepackage{graphicx}
\usepackage{booktabs} % for professional tables

\usepackage{amsmath}
\usepackage{amssymb}
\usepackage{dsfont}

\usepackage[utf8]{inputenc} 
\usepackage[T1]{fontenc}    
\usepackage{url}            
\usepackage{booktabs}       
\usepackage{amsfonts}       
\usepackage{nicefrac}       
\usepackage{microtype}      
\usepackage{graphicx}
\usepackage{mathabx}
\usepackage{longtable}
\usepackage{amsthm}   
\usepackage{stackrel}  

\usepackage{subfig}
\usepackage{tikz,siunitx}
\usetikzlibrary{shapes.geometric,shapes.symbols}

\providecommand{\realnum}{\mathbb{R}}

\providecommand{\bmcal}[1]{\bm{\mathcal{#1}}}

\providecommand{\matnot}[1]{_{[{#1}]}}

\usepackage{bm} 
\usepackage{multirow}
\usepackage{xcolor,colortbl} 
\usepackage{color}

\definecolor{color1}{rgb}{1.0000, 0.1098, 0.0}  
\definecolor{color2}{rgb}{0.8000, 0.0, 0.4000}  
\definecolor{color3}{rgb}{0.0, 1.0000, 0.0}     
\newcommand{\tcolora}[1]{\textcolor{color1}{#1}}
\newcommand{\tcolorb}[1]{\textcolor{color2}{#1}}
\newcommand{\tcolorc}[1]{\textcolor{color3}{#1}}

\usepackage{hyperref}

\usepackage[accepted]{icml2020}

\icmltitlerunning{Multilinear Latent Conditioning for Generating Unseen Attribute Combinations}

\begin{document}

\twocolumn[
\icmltitle{Multilinear Latent Conditioning for Generating Unseen Attribute Combinations}

% It is OKAY to include author information, even for blind
% submissions: the style file will automatically remove it for you
% unless you've provided the [accepted] option to the icml2020
% package.

% List of affiliations: The first argument should be a (short)
% identifier you will use later to specify author affiliations
% Academic affiliations should list Department, University, City, Region, Country
% Industry affiliations should list Company, City, Region, Country

% You can specify symbols, otherwise they are numbered in order.
% Ideally, you should not use this facility. Affiliations will be numbered
% in order of appearance and this is the preferred way.
%\icmlsetsymbol{equal}{*}

\begin{icmlauthorlist}
\icmlauthor{Markos Georgopoulos}{to}
\icmlauthor{Grigorios G Chrysos}{to}
\icmlauthor{Maja Pantic}{to}
\icmlauthor{Yannis Panagakis}{goo}
\end{icmlauthorlist}

\icmlaffiliation{to}{Department of Computing, Imperial College London, United Kingdom}
\icmlaffiliation{goo}{Department of Informatics and Telecommunications, University of Athens, Greece}
\icmlcorrespondingauthor{Markos Georgopoulos}{m.georgopoulos@imperial.ac.uk}
%\icmlcorrespondingauthor{Eee Pppp}{ep@eden.co.uk}
% You may provide any keywords that you
% find helpful for describing your paper; these are used to populate
% the "keywords" metadata in the PDF but will not be shown in the document
\icmlkeywords{Conditional generative models, tensor decomposition, VAE, Tucker decomposition}

\vskip 0.3in
]

% this must go after the closing bracket ] following \twocolumn[ ...

% This command actually creates the footnote in the first column
% listing the affiliations and the copyright notice.
% The command takes one argument, which is text to display at the start of the footnote.
% The \icmlEqualContribution command is standard text for equal contribution.
% Remove it (just {}) if you do not need this facility.

\printAffiliationsAndNotice{}  % leave blank if no need to mention equal contribution
%\printAffiliationsAndNotice{\icmlEqualContribution} % otherwise use the standard text.

\begin{abstract}
Deep generative models rely on their inductive bias to facilitate generalization, especially for problems with high dimensional data, like images. However, empirical studies have shown that variational autoencoders (VAE) and generative adversarial networks (GAN) lack the generalization ability that occurs naturally in human perception.
For example, humans can visualize a woman smiling after only seeing a smiling man. On the contrary, the standard conditional VAE (cVAE) is unable to generate unseen attribute combinations. To this end, we extend cVAE by introducing a multilinear latent conditioning framework that captures the multiplicative interactions between the attributes. We implement two variants of our model and demonstrate their efficacy on MNIST, Fashion-MNIST and CelebA. Altogether, we design a novel conditioning framework that can be used with any architecture to synthesize unseen attribute combinations.
\end{abstract}

\section{Introduction}
\label{sec:introduction}

\begin{figure}[h!]
  \centering

    \includegraphics[width=\linewidth]{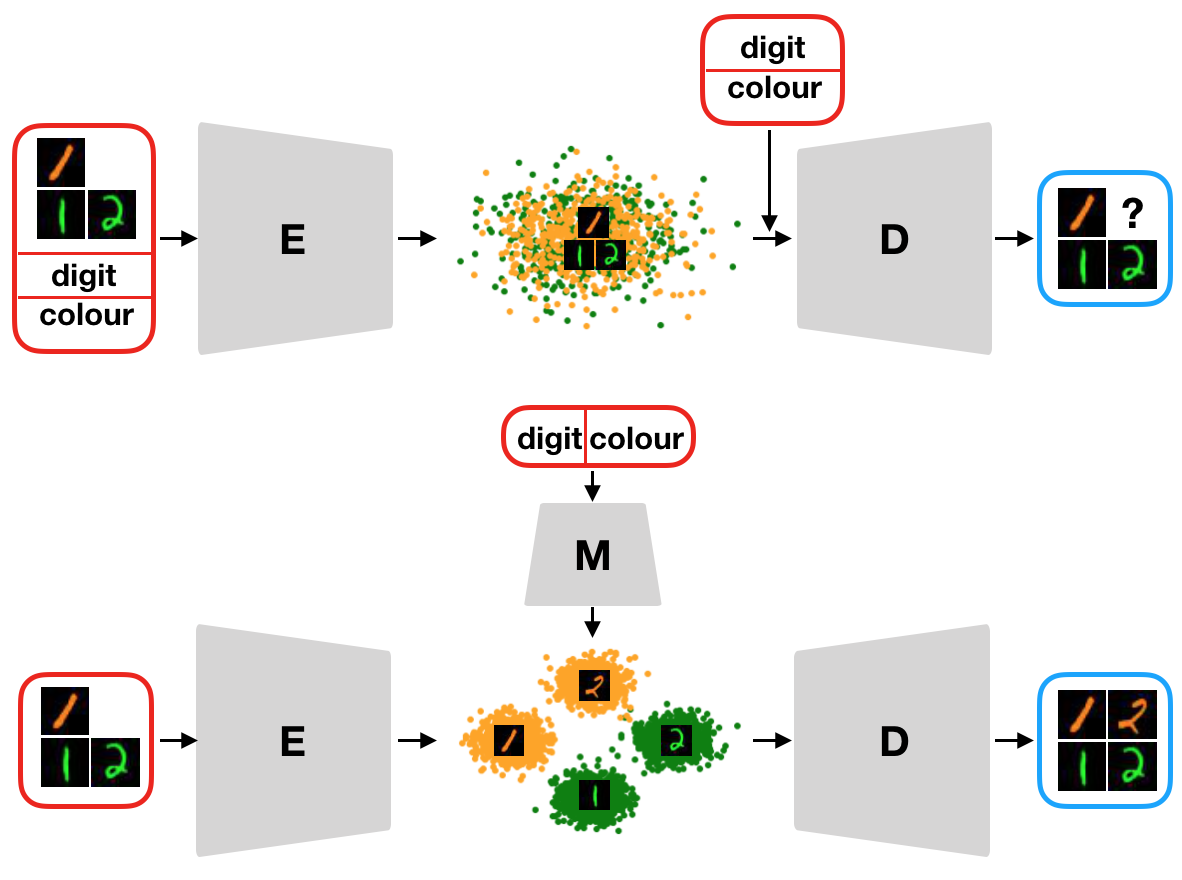}
  \caption{Schematic of the conditional VAE (top) and the proposed framework (bottom). The baseline (conditional VAE) lacks the structure in the latent space and cannot generate images of attribute combinations that are absent from the training set (e.g., orange 2). On the contrary, the proposed method capitalizes on the multilinear latent conditioning to synthesize images of unseen attribute combinations.}
 
  \label{fig:graphical_abstract}
\end{figure}

Deep generative models have been widely utilized to push the boundaries of the state-of-the-art in various tasks including image~\cite{miyato2018spectral}, audio~\cite{andreux2018music}, as well as text synthesis~\cite{radford2017learning}. Their success would not be possible without the development of complex and expressive models \cite{miyato2018spectral, karras2018style, du2019implicit} as well as the collection of rich and thoroughly annotated datasets \cite{liu2015faceattributes,deng2009imagenet}. 
However, due to the high data complexity of these tasks \cite{arora2017gans}, generative models have to rely on a set of assumptions, i.e., their inductive bias.
These assumptions are based on the distribution of the training set, as well as the model itself, and are crucial for generalization.
However, the training set is not always an accurate description of the real world. Hence, the generated distribution is constrained by the distribution of the training data. 

The empirical study conducted in \cite{zhao2018bias} is the first to investigate the inductive bias of deep generative models. The study focuses on the unsupervised setting and demonstrates the inability of variational autoencoders (VAEs) \cite{kingma2013auto,rezende2014stochastic} and generative adversarial networks (GANs) \cite{goodfellow2014generative} to generate images of unseen attribute combinations. For example, consider the task of generating an image of a smiling woman without seeing one in the training set. We posit that in order to further advance the state-of-the-art, generative models should be able to generalize to unseen attribute combinations. In this work, we focus on the supervised setting and investigate the conditional generation of attribute combinations that are absent from the training set.

To this end, using the VAE as the main building block, we propose a multilinear conditioning framework that is tailored to the task. 
The VAE is a generative model that learns the distribution of the data by introducing a latent variable $\bm{z}$. A common assumption is that $\bm{z}$ belongs to a simple prior distribution, for instance, a standard normal. Instead of considering a single prior distribution for all latent representations, we propose to learn a different Gaussian for each class. By learning the mean of each distribution as a multilinear function of the labels, our framework is able to account for their multiplicative interactions. In order to learn the higher order tensors in the presence of missing classes (i.e., attribute combinations), we utilize two tensor decompositions, namely the CP and Tucker, and train the model end-to-end.

We evaluate the efficacy of our method through a series of extensive experiments on benchmarks based on standard datasets. These benchmarks are designed to evaluate both the ability to generate images of unseen attribute combinations, as well as model the correlation between labels. Furthermore, we explore the limits of our model and propose two extensions. Concretely:

\begin{itemize}
    \item In Section \ref{sec:model}, our extension to the VAE and its two variant using the Tucker and CP decompositions are proposed. Contrary to the standard conditional VAE, the proposed framework is able to generate unseen classes. We further exhibit how our model achieves this by modeling the multiplicative interactions between the labels. 
    \item We demonstrate the ability of our models to generalize on unseen attribute combinations in Section \ref{sec:experiments_vae}. In particular, we create 3 evaluation benchmarks based on the MNIST, FashionMNIST and CelebA datasets. For each benchmark, we use 2 sets of labels and remove one label combination from the training set. We evaluate the generated images both qualitatively and quantitatively. The models are also evaluated on attribute transfer on CelebA.
    \item We showcase the importance of modeling inter-label interactions in Section \ref{sec:higher_ord}. To this end, we introduce two benchmarks based on MNIST and CelebA. We show experimentally that only our multilinear latent conditioning framework is able to recover the missing attribute combinations in the presence of highly correlated labels.
    \item In Section \ref{sec:ablation} we explore the proposed model. First, we study how the model performance scales with the number of missing attribute combinations. Furthermore, we extend the framework to handle multiple labelled attributes, as well as unlabelled attributes.
\end{itemize}{}

\section{Related work}

\label{sec:related_vae}
The variational autoencoder is a widely adopted class of generative models. In the seminal work of \citet{kingma2013auto}, VAE learns a generative model of the form $p(\bm{x}, \bm{z}) = p(\bm{x}|\bm{z)}p(\bm{z})$, where $p(\bm{z})$ is the prior distribution of the latent variable.
A common choice for the prior distribution is a standard normal distribution, however this assumption can be relaxed.
More expressive priors have been proposed to model the multimodal nature of the data, e.g., mixture of priors \cite{tomczak2017vae,dilokthanakul2016deep,gronbech2018scvae,jiang2016variational,van2017neural}, normalizing flows \cite{rezende2015variational} and autoregressive models \cite{chen2016variational}. 
The VAE framework was originally proposed for the unsupervised setting but several successful supervised and semi-supervised variations have been proposed thereafter \cite{van2016conditional,razavi2019generating,siddharth2017learning}.

In order to incorporate label information into the model, a common practice is concatenating the labels to the latent representation \cite{sohn2015learning}. However, doing so fails to uncover the underlying structure of the latent space. \citet{klys2018learning} introduce a low-dimensional subspace that captures the class-specific latent structure and use it to perform attribute transfer. Similarly, our model learns the class specific structure of the latent space by considering the multiplicative interactions between the labels.

The multilinear structure of different modes of variation has been previously studied in the unsupervised setting.  \citet{tang2013tensor} extend the Factor Analyzers method to also model multiplicative interactions between the latent factors.  \citet{wang2017disentangling} introduce an unsupervised tensor decomposition that learns the underlying multilinear structure of the data.
In this work, we focus on the supervised setting and propose to model the interactions between the labeled modes of variation.

Lastly, in order to synthesize images of a specified attribute combination, the proposed model implicitly disentangles the labelled information.
Disentanglement of factors of variation is an active research topic in the unsupervised setting \cite{higgins2016beta,kim2018disentangling,chen2018isolating,burgess2018understanding,ridgeway2018learning,siddharth2017learning}. \citet{locatello2019challenging} proved that unsupervised learning of disentangled representations is not possible without inductive biases.
Furthermore, recent work \cite{khemakhem2020variational} highlights the lack of identifiability without auxiliary information, such as labels. 
In this work, we focus on the supervised setting and aim to recover the unobserved attribute combinations, a task that is not a direct byproduct of disentanglement.

Concurrently with this work, \cite{bozkurt2019evaluating} evaluate whether VAEs can reconstruct data of unseen classes. This work differentiates itself by focusing on the generation of samples of unobserved attribute combinations.

\section{Background}
\label{sec:background}

\subsection{Notation}
Tensors are symbolized by calligraphic letters, e.g.,  $\bmcal{X}$, while matrices (vectors) are denoted by uppercase (lowercase) boldface letters e.g., $\bm{X}$, ($\bm{x}$). 

\textbf{Products}: 
The \textit{Hadamard} product of $\bm{A} \in \realnum^{I \times N}$ and $\bm{B} \in \realnum^{I \times N}$ is defined as $\bm{A} * \bm{B}$ and is equal to ${A}_{(i, j)} {B}_{(i, j)}$ for the $(i, j)$ element. 

The \textit{Kronecker} product product of matrices $\bm{A} \in \realnum^{I \times M}$
and $\bm{B} \in \realnum^{J \times N}$ is
denoted by $\bm{A} \otimes \bm{B}$ and yields a matrix of dimensions $(IJ)\times (MN)$ that is defined as: 
\begin{align}
\bm{A} \otimes \bm{B} &= 
\begin{bmatrix} 
    a_{11}\bm{B} & a_{12}\bm{B} & \dots & a_{1J}\bm{B} \\
    a_{21}\bm{B} & a_{22}\bm{B} & \dots & a_{2J}\bm{B} \\
    \vdots & \vdots & \ddots & \vdots \\
    a_{I1}\bm{B} & a_{I2}\bm{B} & \dots & a_{IJ}\bm{B}
\end{bmatrix}.
\end{align}

The \textit{Khatri-Rao} (column-wise \textit{Kronecker}) product of matrices $\bm{A} \in \realnum^{I \times N}$
and $\bm{B} \in \realnum^{J \times N}$ is
denoted by $\bm{A} \odot \bm{B}$ and yields a matrix of
dimensions $(IJ)\times N$ that is defined as:
\begin{align}
    \bm{A} \odot \bm{B} =
    \begin{bmatrix} \bm{a}_1 \otimes \bm{b}_1 & \bm{a}_2 \otimes \bm{b}_2 & \cdots & \bm{a}_N \otimes \bm{b}_N \end{bmatrix}.
\end{align}
In the case of vectors, the Khatri-Rao and Kronecker products are equivalent.

\subsection{Tensor operations and decompositions}

Each element of a $K^{th}$ order tensor $\bmcal{X}$ is addressed by $K$ indices, i.e., $(\bmcal{X})_{i_{1}, i_{2}, \ldots, i_{K}} \doteq x_{i_{1}, i_{2}, \ldots, i_{K}}$. A $K^{th}$-order real-valued tensor $\bmcal{X}$ is  defined over the
tensor space $\realnum^{I_{1} \times I_{2} \times \cdots \times I_{K}}$, where $I_{k} \in \mathbb{Z}$ for $k=1,2,\ldots,K$.

\textbf{Tensor unfolding}: The \textit{mode-$k$ unfolding} of a tensor $\bmcal{X} \in
 \realnum^{I_1 \times I_2 \times \cdots \times I_K}$ maps
 $\bmcal{X}$ to a matrix $\bm{X}_{(k)} \in \realnum^{I_{k}
 \times \bar{I}_{k}}$ with $\bar{I}_{k}= \prod_{t=1 \atop t  \neq k}^K I_t $ such
 that the tensor element $x_{i_1, i_2, \ldots, i_K}$ is
 mapped to the matrix element $x_{i_{t}, j}$ where
 $j=1 + \sum_{t=1 \atop t \neq k}^K (i_t - 1) J_t$ with $J_t =
\prod_{n =1 \atop n \neq k}^{t-1} I_n $.

\textbf{Vector product}: The \textit{mode-$k$ vector product} of $\bmcal{X}$ with a
vector $\bm{u} \in \realnum^{I_k}$, denoted by
$\bmcal{X} \times_{k} \bm{u} \in \realnum^{I_{1}\times
I_{2}\times\cdots\times I_{k-1}  \times I_{k+1} \times
\cdots \times I_{K}} $, results in a tensor of order $K-1$:
\begin{equation}\label{E:Tensor_Mode_n}
(\bmcal{X} \times_{k} \bm{u})_{i_1, \ldots, i_{k-1}, i_{k+1},
\ldots, i_{K}} = \sum_{i_k=1}^{I_k} x_{i_1, i_2, \ldots, i_{K}} u_{i_k}.
\end{equation}

\textbf{CP tensor decomposition}:
The \textit{CANDECOMP/PARAFAC (CP) decomposition}~\citep{kolda2009tensor,Sidiropoulos:16} factorizes a tensor into a sum of component rank-one tensors. The rank-$R$ CP decomposition of a $K^{th}$-order tensor $\bmcal{X}$ is written as:
 \begin{equation}\label{E:CP}
\bmcal{X}  \doteq [\![ \bm{U}\matnot{1}, \bm{U}\matnot{2}, \ldots, \bm{U}\matnot{K}  ]\!] =  \sum_{r=1}^R \bm{u}_r^{(1)}  \circ \bm{u}_r^{(2)}  \circ \cdots \circ \bm{u}_r^{(K)},
\end{equation}
where $\circ$ denotes for the vector outer product. The factor matrices $\big\{ \bm{U}\matnot{k} = [\bm{u}_1^{(k)},\bm{u}_2^{(k)}, \cdots, \bm{u}_R^{(k)} ] \in \realnum^{I_k \times R} \big\}_{k=1}^{K}$ collect
the vectors from the rank-one components. By considering the mode-$1$ unfolding of $\bmcal{X}$, the CP decomposition of a third order tensor can be written in matrix form as \citep{kolda2009tensor}:
 \begin{equation}
 \label{eq:cp_unfolding}
\bm{X}_{(1)}  
\doteq \bm{U}\matnot{1} (\bm{U}\matnot{3}\odot \bm{U}\matnot{2})^T.
\end{equation}

\textbf{Tucker tensor decomposition}: The Tucker tensor decomposition \cite{tucker1963implications} factorizes a tensor into a set of matrices and a core tensor, as follows:
\begin{align}\label{E:Tucker}
\bmcal{X}  &\doteq [\![ \bm{\mathcal{G}}; \bm{U}\matnot{1}, \bm{U}\matnot{2}, \ldots, \bm{U}\matnot{K}  ]\!]  \nonumber\\
 &= \bm{\mathcal{G}}  \times_1 \bm{U}\matnot{1} \times_2 \bm{U}\matnot{2}\times_3 \dots \times_K \bm{U}\matnot{K},
\end{align}
Similarly to CP , the Tucker decomposition of a third order tensor can be written in matrix form as: 
\begin{equation}
 \label{eq:tucker_unfolding}
\bm{X}_{(1)}  
%\doteq \bm{U}\matnot{1} \bm{G}_{(1)} \bigg( \bigotimes_{k = K}^{2} \bm{U}\matnot{k}\bigg)^T
\doteq \bm{U}\matnot{1} \bm{G}_{(1)} (\bm{U}\matnot{3}\otimes \bm{U}\matnot{2})^T.
\end{equation}
Tucker can be considered a generalization of SVD to higher order tensors \cite{de2000multilinear}. The CP decomposition is a special case of Tucker, where the core tensor is superdiagonal.

 \section{Model}
\label{sec:model}

Consider the setting of a dataset of images $\bm{x}$ that are annotated with a set of $N$ one-hot label vectors $ \bm{y} = \{ \bm {y}_{i} \in \realnum^{D_i} \}_{i=1}^N$ and latent representations $\bm{z}\in \realnum^{d}$. The conditional VAE (cVAE) \cite{sohn2015learning} is trained by maximizing the following objective:
\begin{align}
L_{CVAE} =&  \mathbb{E}_{q_{\phi}(\bm{z}|\bm{x}, \bm{y})}\left[\log p_\theta(\bm{x} | \bm{z}, \bm{y})\right] \nonumber\\
&- \beta D_{KL}\left( q_\phi(\bm{z}|\bm{x}, \bm{y})\parallel p(\bm{z}) \right) \leq \log p_\theta(\bm{x} | \bm{y}),
\end{align}
where $D_{KL}$ is the Kullback Leibler divergence between the approximate posterior $q_{\phi}$ and the prior $p(\bm{z})$. The approximate posterior is modeled by the encoder network, while the prior is assumed to be standard normal.

In this work, we introduce Multilinear Conditioning VAE (MLC-VAE), which is a generative model of the form $p(\bm{x}, \bm{y}, \bm{z}) = p(\bm{x}|\bm{z)}p(\bm{y}|\bm{z})p(\bm{z})$. We modify the objective of cVAE by introducing a label reconstruction term $\log p_{\theta_{\bm{y}}}(\bm{y} | \bm{z})$ which is parametrized by a label decoder. The new evidence lower bound (ELBO) is of the form:
\begin{align}
L_{MLC-VAE} = &\mathbb{E}_{q_{\phi}(\bm{z}|\bm{x}, \bm{y})}\left[
\log p_{\theta_{\bm{x}}}(\bm{x} | \bm{z}) + 
\log p_{\theta_{\bm{y}}}(\bm{y} | \bm{z})\right] \nonumber\\
&- \beta D_{KL}\left(  q_{\phi}(\bm{z}|\bm{x}, \bm{y}) \parallel p(\bm{z}) \right)
\label{E:elbo}
\end{align}
Instead of assuming a single prior distribution, we consider that the latent representations of each class (i.e., combination of attribute labels $\bm{y}_1,\dots, \bm{y}_N$) are centered around a different mean, that is, $\bm{z}\sim \mathcal{N}(\bm{\mu}_{y_1,\dots, y_N}, \bm{I})$. The class means $\bm{\mu}_{y_1,\dots, y_N} \in \realnum^{d}$ are learned as a multivariate function $M(\bm{y}_1, \dots, \bm{y}_N)$ of the label vectors. Thus, we are able to obtain a latent representation for every attribute combination, even if it is not represented in the training set.

A simple choice for $M$ would be the linear combination of each attribute. Concretely, in the case of the 2 attributes $\bm{y}_1$ and $\bm{y}_2$:
\begin{equation}
    M(\bm{y}_1, \bm{y}_2) = \bm{W}^{[1]} \bm{y}_1 + \bm{W}^{[2]} \bm{y}_2
    \label{E:lcvae}
\end{equation}
However, this formulation completely ignores the non-linear (e.g., multiplicative) interactions between the labels. To account for the multiplicative interactions, we introduce a higher order term that models the correlations between the attributes:
\begin{equation}
    M(\bm{y}_1, \bm{y}_2) =\bm{W}^{[1]} \bm{y}_1+  \bm{W}^{[2]} \bm{y}_2+\bm{\mathcal{W}}^{[12]} \times_2 \bm{y}_1 \times_3 \bm{y}_2
    \label{E:model_2_atts}
\end{equation}
The first two terms in \ref{E:model_2_atts} are equal to the linear combination of the attribute labels, with the matrices $\bm{W}^{[1]}$ and $\bm{W}^{[2]}$ learning a mapping between each individual label and the latent space. The third order tensor $\bm{\mathcal{W}}^{[12]}$ captures the pairwise multiplicative interaction between $\bm{y}_1$ and $\bm{y}_2$ and learns a latent representation for each attribute combination. We investigate the effect of the higher order term in Section \ref{sec:higher_ord}. 

We can rewrite the higher order term in \ref{E:model_2_atts} by replacing the tensor by its \textit{mode-1} unfolding as such:  
\begin{align}
M(\bm{y}_1, \bm{y}_2) 
=&\bm{W}^{[1]} \bm{y}_1+  \bm{W}^{[2]} \bm{y}_2+\bm{W}_{(1)}^{[12]} (\bm{y}_2 \odot \bm{y}_1)
\label{2_att_poly}
\end{align}

Since $\bm{y}_1 \in \realnum^{D_1}$ and $\bm{y}_2 \in \realnum^{D_2}$ are one-hot encoded vectors, their Khatri-Rao product $(\bm{y}_2 \odot \bm{y}_1)$ yields a new one-hot vector $\bm{y} \in \realnum^{D_1  D_2}$. Therefore, the product $\bm{W}_{(1)}^{[12]} (\bm{y}_2 \odot \bm{y}_1)$ yields the column of the matrix $\bm{W}_{(1)}^{[12]}$ that corresponds to class ($\bm{y}_1, \bm{y}_2$). Hence, in the presence of absent attribute combinations, it is not possible to learn the matrix using stochastic gradient descent based optimizers. However, if we let the tensor $\bm{\mathcal{W}}^{[12]}$ admit a low-rank decomposition, we can overcome this issue as we show below. We explore the \textit{CP} and \textit{Tucker} decompositions and propose two different models, namely MLC-VAE-T and MLC-VAE-CP. 

After applying the CP and Tucker decompositions on \ref{2_att_poly}, we obtain respectively:
\begin{align}
&M_{CP} =\bm{W}^{[1]} \bm{y}_1+  \bm{W}^{[2]} \bm{y}_2+
\bm{U}_{[1]} (\bm{U}_{[3]}^T\bm{y}_2 * \bm{U}_{[2]}^T\bm{y}_1)
\label{E:MCP}
\end{align}
\begin{align}
&M_{T} =\bm{W}^{[1]} \bm{y}_1+  \bm{W}^{[2]} \bm{y}_2+
\bm{U}_{[1]} \bm{G}_{(1)}^{[12]} (\bm{U}_{[3]}^T\bm{y}_2 \otimes \bm{U}_{[2]}^T\bm{y}_1) 
\label{E:MT}
\end{align}

With this formulation, we can obtain $\mu_{y_1, y_2}$ for every class combination $(\bm{y_1}, \bm{y_2})$ even when not all possible combinations are seen during training. Thereafter, we can optimize \ref{E:elbo} for $\bm{z}\sim \mathcal{N}(M(\bm{y_1}, \bm{y_2}), \bm{I})$.

\begin{figure*}[htb] 
    \centering
        \includegraphics[width=\textwidth]{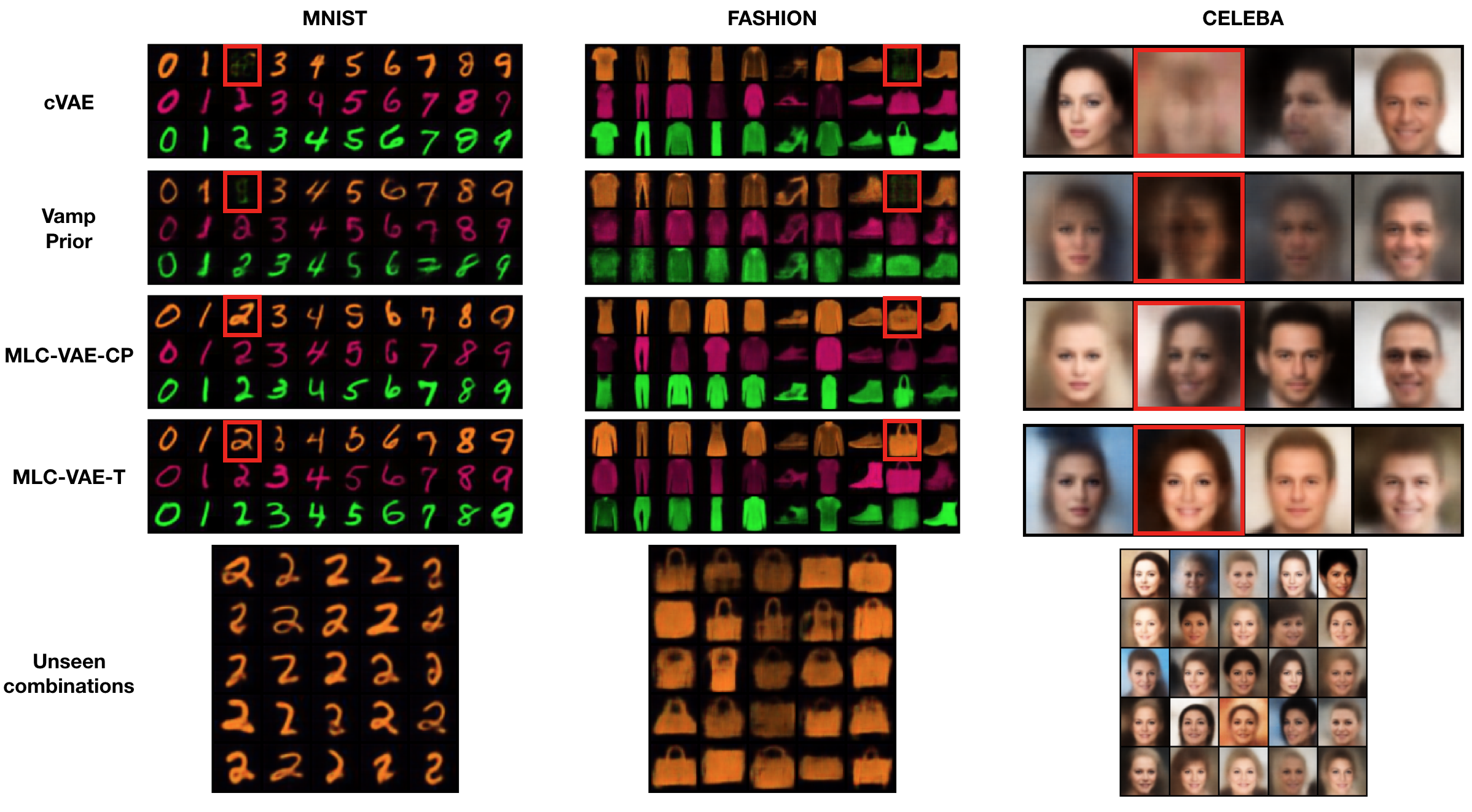}
    \caption{(top) Samples synthesized by the compared methods on the MNIST, Fashion-MNIST, CelebA. All methods can synthesize samples from combinations that are present in the training set, however the compared methods fail on unseen combinations (marked with red boxes). (bottom) Synthesized samples corresponding to combinations that are not seen during training (orange 2, orange bag and woman with smile). Our method is able to synthesize diverse samples of these combinations.}
    \label{fig:results}
\end{figure*} 
 \section{Experiments}
\label{sec:experiments_vae}

To evaluate our model on multi-attribute conditional image generation, we perform experiments on the MNIST~\cite{lecun1998gradient}, Fashion-MNIST \cite{xiao2017fashion} and CelebA \cite{liu2015faceattributes} datasets. In all the experiments the decoder networks are convolutional architectures based on DCGAN \cite{radford2015unsupervised}. The encoders are constructed as the 'mirrored' versions of the decoders. We compare both proposed methods with the baseline cVAE \cite{sohn2015learning} and VampPrior \cite{tomczak2017vae}; we focus on generation of the unseen combinations but also visualize samples from the seen classes. To evaluate the models quantitatively, we train attribute classifiers. In particular, we train a convolutional neural network with the same architecture as the corresponding VAE encoder for all attributes except for color. For the color attribute used in MNIST and Fashion-MNIST, we train an MLP on the color histograms of the images. The classification accuracy on the test set was above 97\% for all attribute classifiers, except for smile, where the accuracy was 92\%. The quantitative results on the unseen attribute combinations are presented in Table \ref{tab:quant_results}. 

Overall, the results in Table \ref{tab:quant_results} highlight the inability of the baseline models to generate novel attribute combinations. The accuracy of the classifiers on the unseen combinations are mostly worse than random chance. This implies that, contrary to our models, the baselines are not able to synthesize recognizable digits, objects and faces of the target attributes. This is verified qualitatively in Figure \ref{fig:results}. On the other hand, all models are able to generate the seen attribute combinations (Figure \ref{fig:results}).
Quantitative results for the seen attribute combinations can be found in the supplementary material.

\textbf{Implementation details}:  For the experiments on MNIST and Fashion-MNIST the encoder and decoder networks have 4 layers, while the networks for CelebA have 5 layers. All label decoders are affine transformations. We set $\beta=1$ for all experiments, except for CelebA where we set $\beta=10$. The input label for the cVAE and VampPrior baselines is a one-hot vector where each class corresponds to an attribute combination (e.g., (1, orange) or (Male, Smiling)). For fair comparison we train all models for
50 epochs using the Adam optimizer \cite{kingma2014adam} with a learning rate of $0.0005$. All models were implemented in Pytorch~\cite{paszke2017automatic} and Tensorly \cite{kossaifi2019tensorly}.

\textbf{MNIST}: The MNIST dataset consists of 60k training images and 10k test images of handwritten digits. The original images are greyscale. Since the dataset is only annotated with regards to one attribute (digit) we introduce a second mode of variation and corresponding labels. That is, we color each digit using 3 different colors, namely orange, purple and green. We then remove from the training set the images with labels (2, orange). The quantitative results for the unseen combination are reported in Table \ref{tab:quant_results}, while the qualitative in Figure \ref{fig:results}. The proposed models are able to synthesize the unseen attribute combination and, as seen in the figure, they demonstrate considerable variation in the synthesized styles of the unseen combination.

\begin{figure}[!h]
  \centering 
    \includegraphics[width=0.8\linewidth]{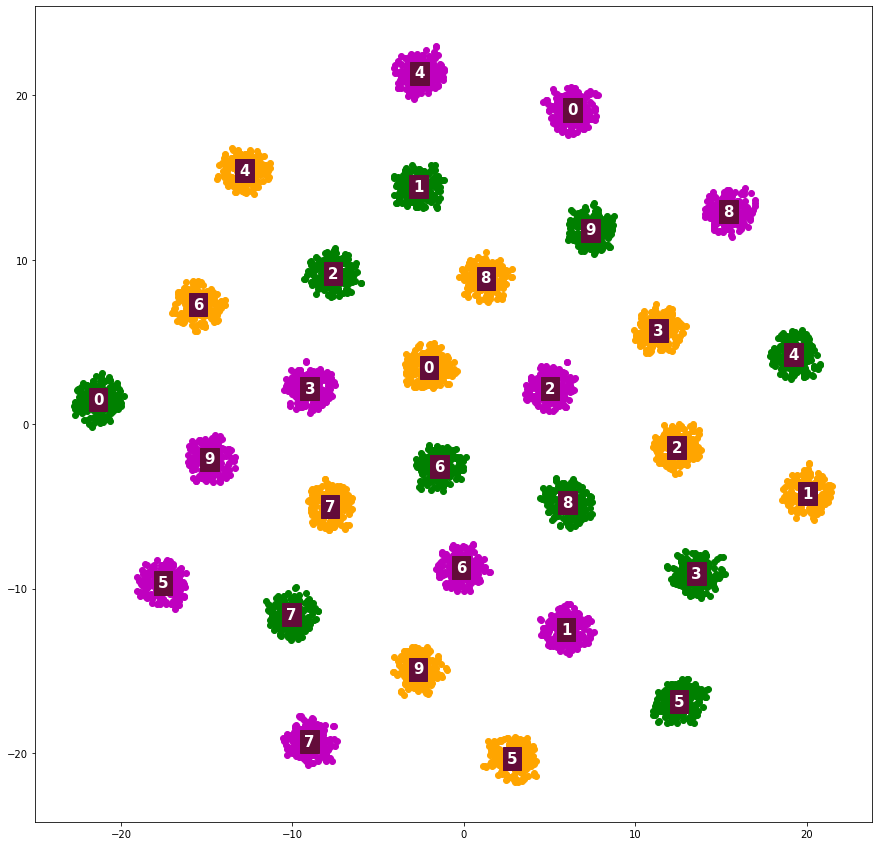}
  \caption{T-SNE~\cite{maaten2008visualizing} on the latent space of the proposed model during prediction time. Note that without explicit constraints, the latent space is naturally clustered based on the attribute labels.}
  \label{fig:tsne_mnist}
\end{figure}

\begin{figure}[!b] 
    \centering

    \subfloat[Interpolation \tcolorb{6} $\rightarrow$ \tcolora{6} $\rightarrow$ \tcolorc{6}]{
        \includegraphics[width=1\linewidth]{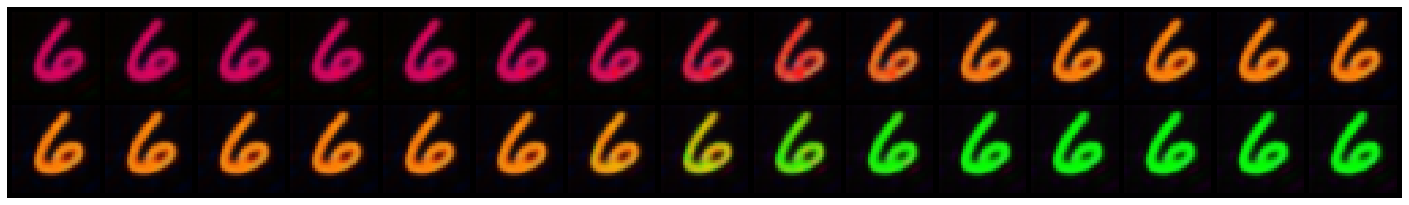}
        \label{fig:interpolation_mnist}
        }
    \hfill
    \subfloat[Interpolation WNS $\rightarrow$ WS  and MNS $\rightarrow$ MS]{
        \includegraphics[width=1\linewidth]{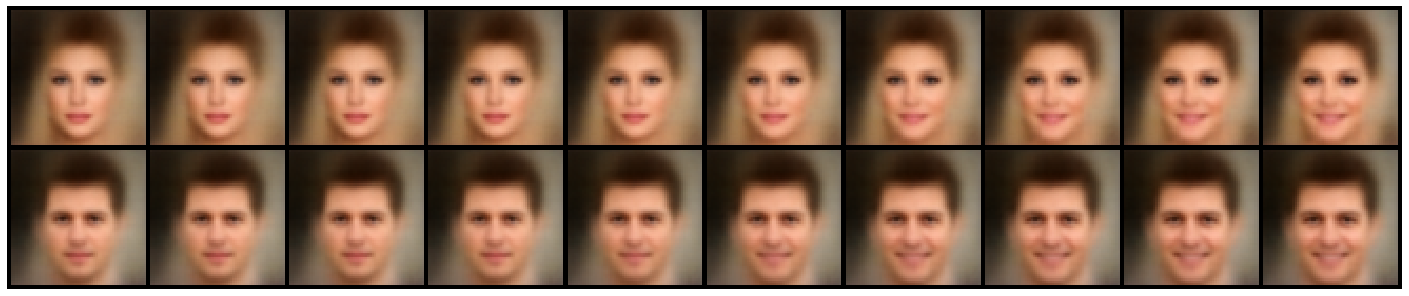}
        \label{fig:interpolation_celeba}
        }
        
    \caption{Linear interpolation results on (a) MNIST, (b) CelebA. Specifically, in (a) the interpolation is on the color attribute while in (b) the interpolation is across the gender and the smiling attribute. Note that in (a) unseen colors emerge. }
    \label{fig:interpolations}
\end{figure}

The latent space of the model is projected into two-dimensions with T-SNE and visualized in Figure \ref{fig:tsne_mnist}. The model disentangles the digit and color information and creates class-specific clusters. As customary in image generation, linear interpolations (in the color attribute) are depicted in Figure \ref{fig:interpolation_mnist}.

\begin{figure}[!b] 
    \centering

    \subfloat[Transfer of smile and gender]{\includegraphics[width=0.44\linewidth]{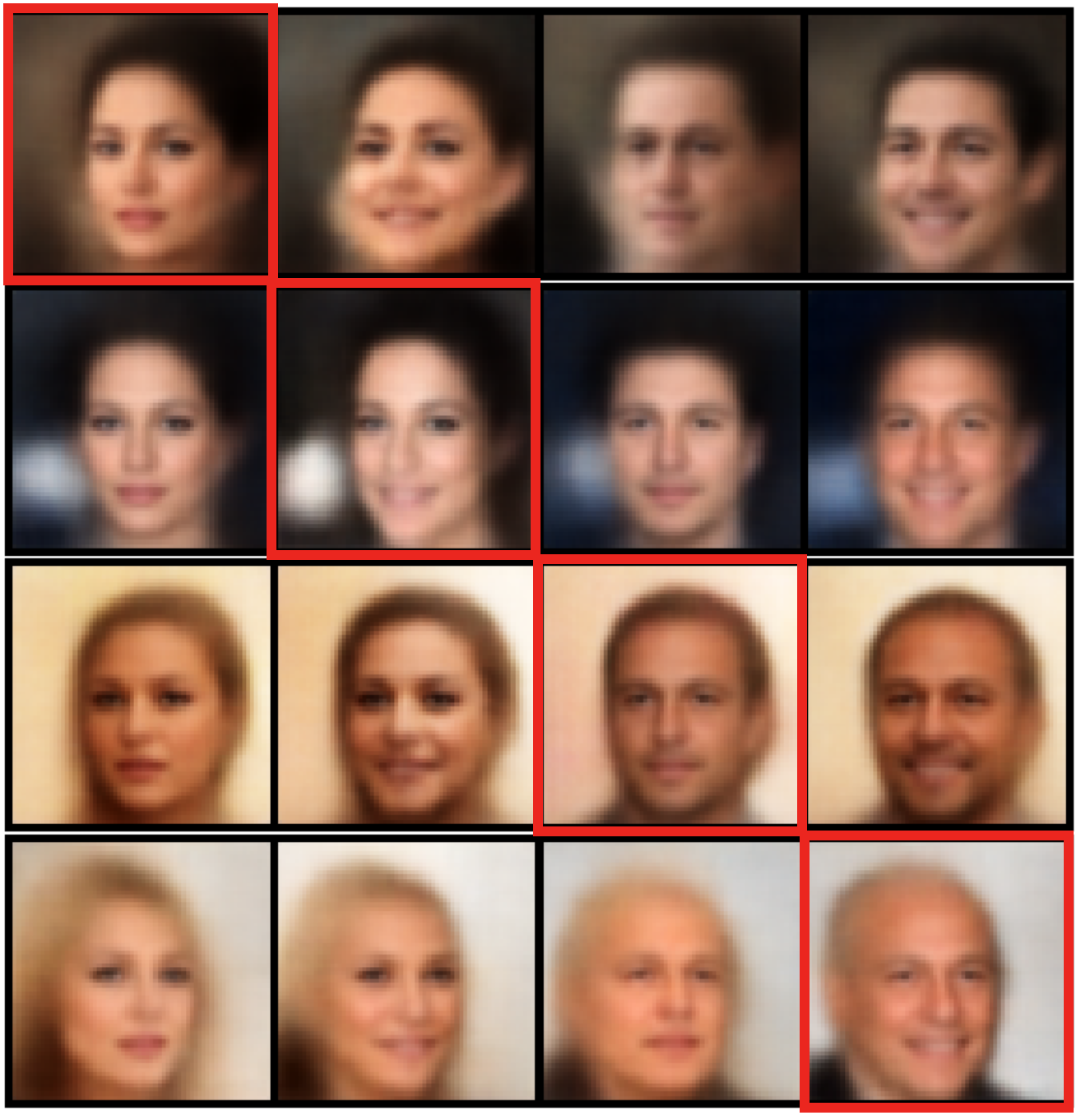}
        \label{fig:smile_transfer}
        }
    \hfill
    \subfloat[Synthesized samples for the ("Male", "Eyeglasses") protocol]{\includegraphics[width=0.44\linewidth]{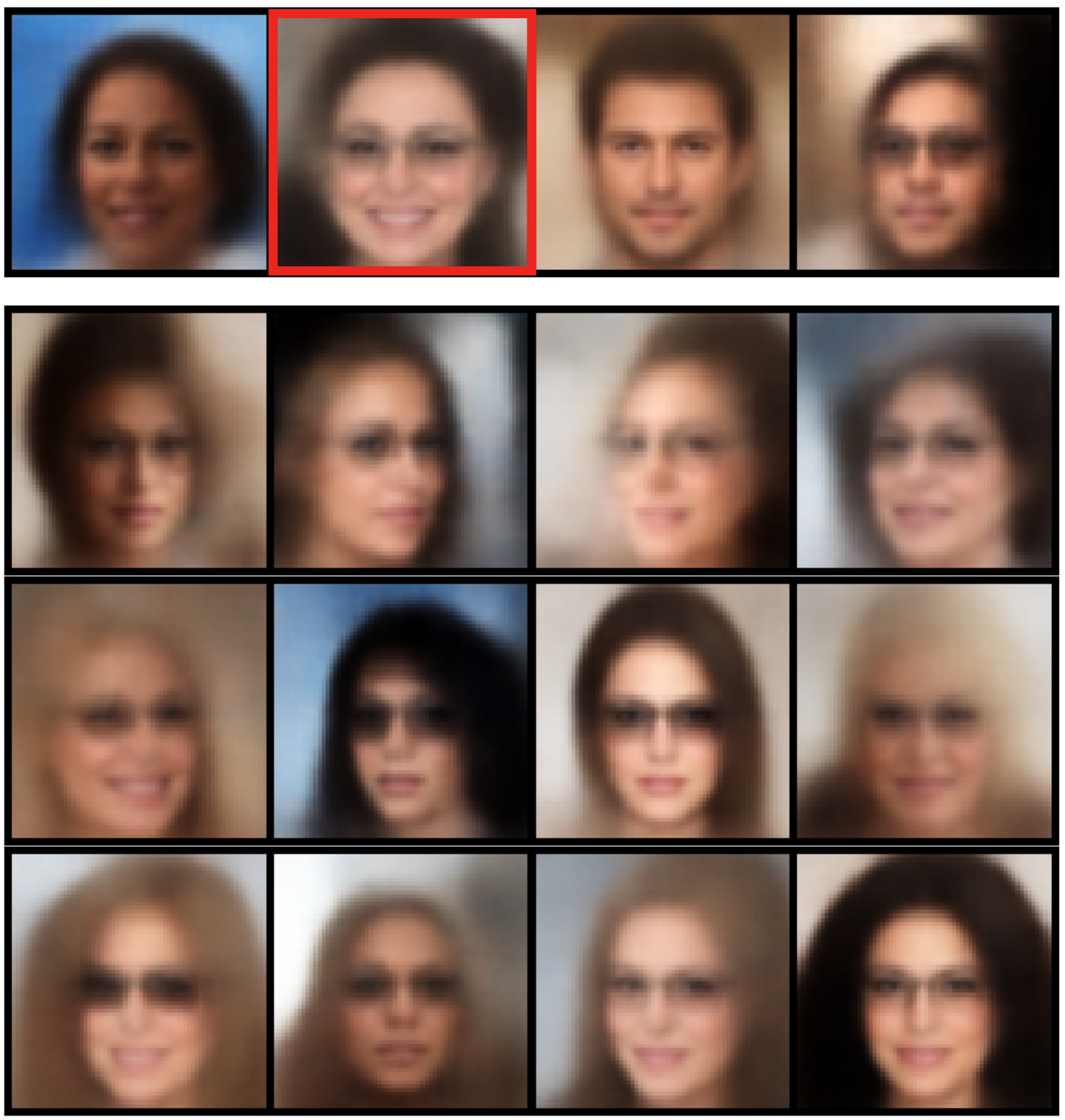}
        \label{fig:celaba_glasses}
        }
        
    \caption{(a) Transfer of attributes in CelebA. The faces in the red rectangle are the original ones. For each row, we translate the image in the red square into the 3 remaining classes. (b) Synthesized samples for attributes gender and wearing glasses. The first row contains generated samples from all the classes (missing class in the red square). The bottom 3 rows contain generated samples just from the unseen class (woman with eyeglasses).}
    \label{fig:more_celeba_results}
\end{figure}

\textbf{Fashion-MNIST}: Similarly to MNIST, Fashion-MNIST consists of 60k training and 10k test images. We perform the same preprocessing as for MNIST and color each object orange, purple and green. We remove the images with labels (bag, orange) and present qualitative and quantitative results in Figure \ref{fig:results} and Table \ref{tab:quant_results}. Although the baselines fail to generate even the correct color (orange), we see that the proposed models can synthesize the missing combination successfully.

\textbf{CelebA}: The CelebA dataset contains over 200k facial images, annotated with 40 attributes. For our experiments, we chose two of the most balanced attributes, namely "Male" and "Smiling". Thereafter, we remove the images labeled (not "Male", "Smiling"). The images were resized to 64x64.

The results in Table~\ref{tab:quant_results} and Figure~\ref{fig:results} indicate that, contrary to the baselines, the proposed models are able to generate diverse images of smiling women. To ensure that the model generalizes on imbalanced sets of attributes we conduct an experiment with the same hyperparameters using the ("Male", "Eyeglasses") attributes of the dataset. We note that the attribute "Eyeglasses" appears on less than 7\% of the images. Further qualitative results on the ("Male", "Eyeglasses") protocol are presented in Figure \ref{fig:celaba_glasses}. For this experiment we remove the images with labels (not "Male", "Eyeglasses"). Interpolations between attributes are presented in Figure \ref{fig:interpolation_celeba}.

\begin{table*}[hbt!]
\centering    

\begin{tabular}{ccccccccc}

\hline
\textbf{Model}&\multicolumn{2}{c}{\textbf{MNIST}}& \multicolumn{2}{c}{\textbf{FASHION}}& \multicolumn{2}{c}{\textbf{CELEB-A}}&\multicolumn{2}{c}{\textbf{CELEB-A Transfer}}\\

&digit & color & object & color &gender&smile&gender&smile \\
\hline
Random chance &10&33.3&10&33.3&50&50&50&50\\
\hline
cVAE & 11.5 & 49 & 13.9 & 23 & 18 & 7.6 & 23.1&9\\
VampPrior&2.6&2.3&4&1.8&17.1&2.8&44.3&5.8 \\
\hline
MLC-VAE-CP (Ours)&68.1&99.2&70.3&99.7&96.4&\textbf{94}&95&90.6 \\
MLC-VAE-Tucker (Ours)&\textbf{95.1}&\textbf{100}&\textbf{84.6}&\textbf{99.8}&\textbf{99.4}&93.5&\textbf{99.4}&\textbf{91.5} \\
\hline

\end{tabular}
\centering    
\caption{Accuracy (\%) of the attribute classifier for each model on the \textbf{unseen} attribute combinations of MNIST, FASHION, CELEB-A generation and CELEB-A transfer benchmarks. The proposed models outperform the baselines by a wide margin in all cases. The proposed model performs better overall when combined with the Tucker decomposition compared to CP. Random chance assumes a uniform selection from each attribute. The accuracy of the classifier on seen combinations can be found in the supplementary material.}

\label{tab:quant_results}
\end{table*} 
Besides image synthesis, we evaluate the models on attribute transfer on CelebA. In this setting, given an input image and a set of target attributes $(\bm{y}_1, \bm{y}_2)$, we modify the input image so that its attributes match $\bm{y}_1$ and $\bm{y}_2$. Specifically, we encode an image and then decode it using different target attributes. To evaluate the attribute transfer, we use the attribute classifiers as described above. Quantitative and qualitative results on smile and gender transfer are presented in Table \ref{tab:quant_results} and Figure \ref{fig:smile_transfer}.

 \section{The effect of modeling label interactions}
\label{sec:higher_ord}
In this section we experiment with the choice of the function $M$. In particular we investigate the importance of modeling the label interactions. To this end, we introduce a simpler version of the model that learns the mean as a linear combination of the labels as described in \ref{E:lcvae}.
% \begin{align}{}
% M(\bm{y}_1, \dots, \bm{y}_N) =\sum_{i=1}^{N} \bm{W}^{[i]} \bm{y}_i.
% \end{align}
We label this model LC-VAE. Hereafter, we compare LC-VAE to MLC-VAE on two custom benchmarks.

\textbf{mixed-up MNIST}: The first benchmark is created by introducing correlation between the labels of MNIST. We begin by coloring the MNIST dataset red and blue, as discussed in the previous experiments. Thereafter, we change the color label of each image if the digit is odd. The images of even digits kept their original color labels. The resulting dataset presents clear interaction between the digit and color labels. Similarly to the previous benchmarks, we remove the combination (2, red) from the training set.

\begin{figure}[!h] 
    \centering

    \includegraphics[width=0.9\linewidth]{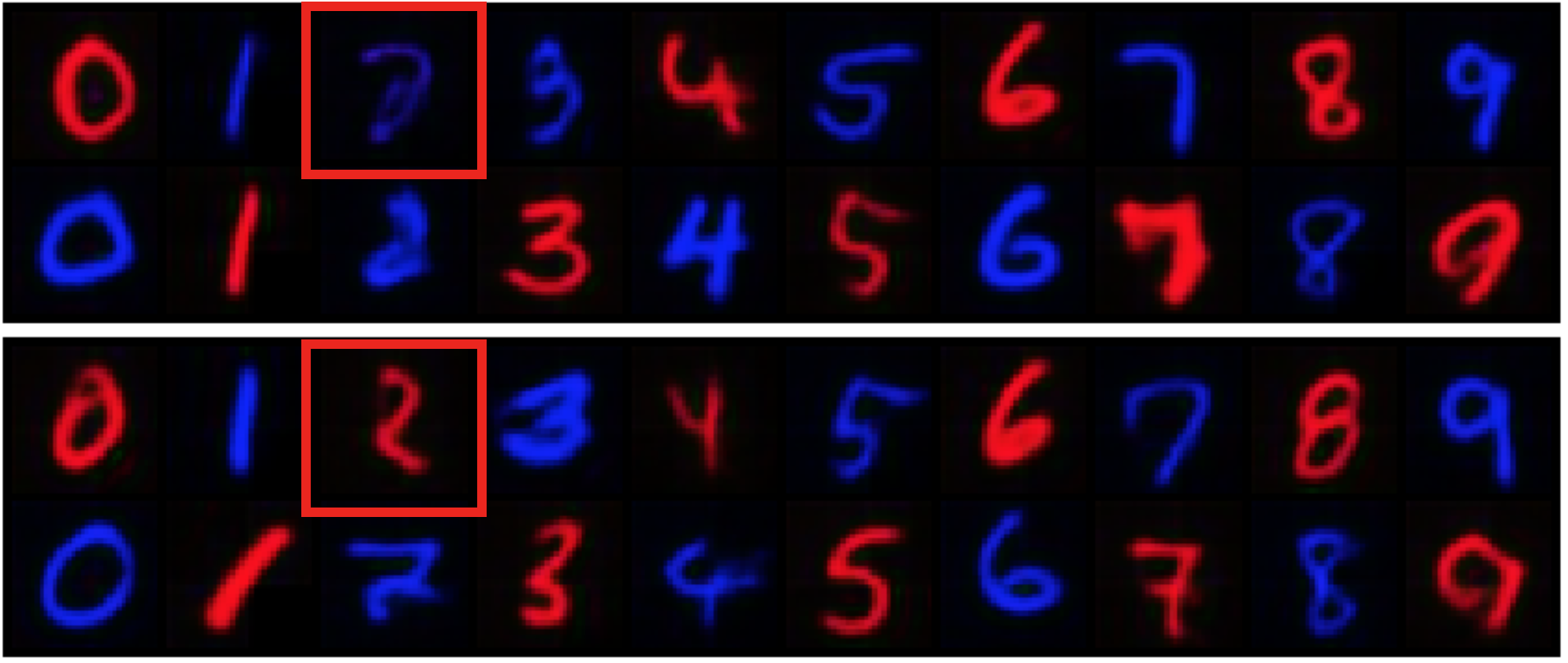}

    \caption{Results on the mixed-up MNIST. The top two rows are synthesized from LC-VAE while the bottom two rows from MLC-VAE. The first and third rows are the digits labelled red and the second and fourth rows are digits labeled blue. We observe that only the MLC-VAE is able to generate the missing 2-red (in the red square).}
    \label{fig:mnist_higher_ord}
\end{figure}

The models are trained as in Section~\ref{sec:model}.
The correlations between the labels, i.e., the color cannot be generated without explicit knowledge of the digit, is not captured by LC-VAE as demonstrated on Figure \ref{fig:mnist_higher_ord}. On the contrary, the proposed models rely on the higher-order terms to capture such multiplicative interactions.

\textbf{CelebA}: We conduct a similar experiment on faces, where such label correlations occur naturally. To do that, we select the "Smiling" and "Mouth Slightly Open" attributes of CelebA. The interaction between the two labels is obvious, as most smiles in CelebA are labelled as "Mouth Slightly Open". We test the models on their ability to generate a smile with the mouth closed by excluding the combination ("Smiling", not "Mouth Slightly Open") from the training set. 

\begin{figure}[h!] 
    \centering
    \includegraphics[width=0.7\linewidth]{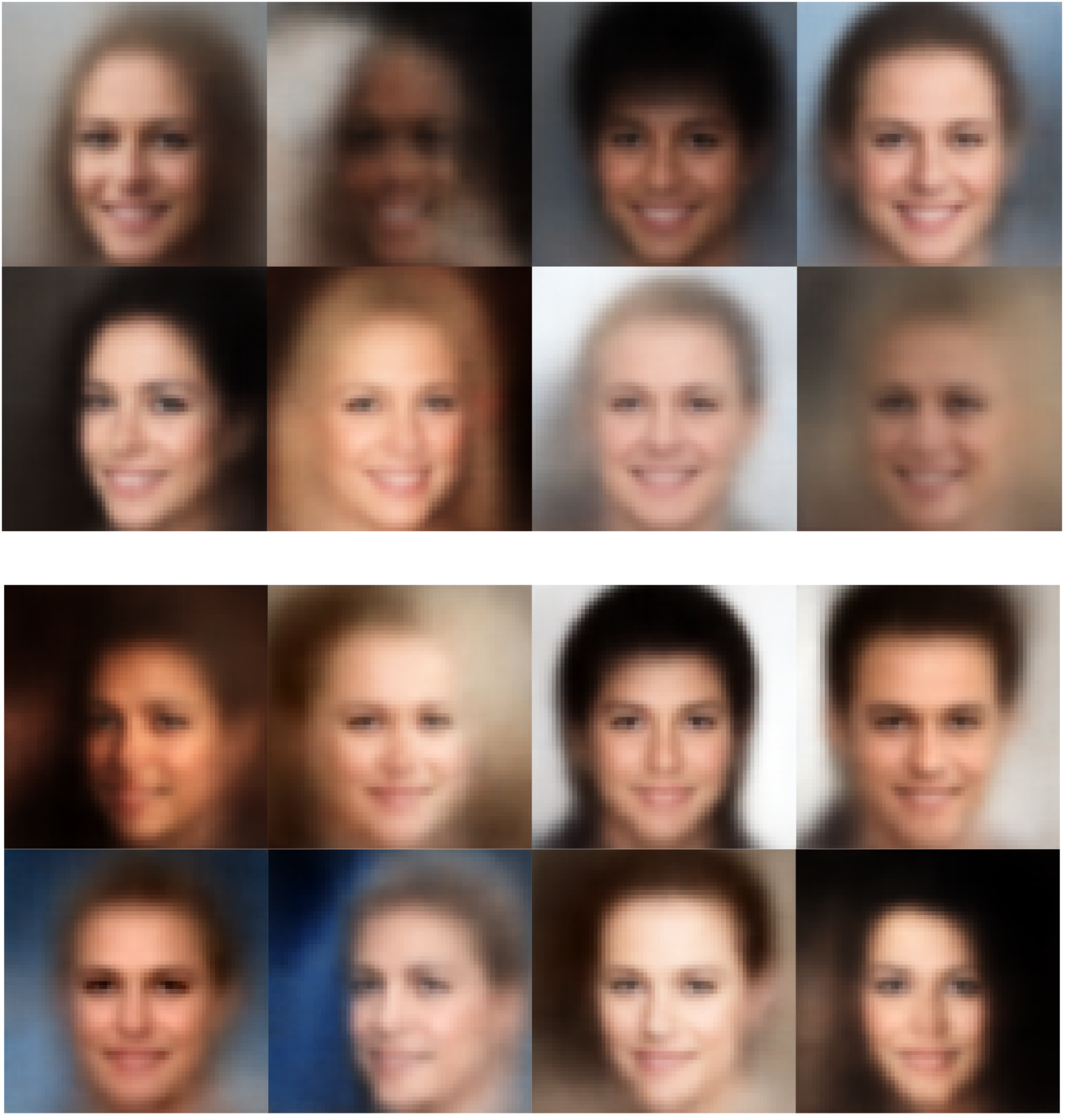}
    \caption{Results on the ("Smiling",  "Mouth Slightly Open") protocol of CelebA. The top two rows are synthesized from LC-VAE while the bottom two rows from MLC-VAE. We observe that only the latter is able to generate the missing combination (Smiling, Closed mouth), while the former ends up generating faces with an open mouth.}
    \label{fig:smile_higher_ord}
\end{figure}

The visual results on Figure \ref{fig:smile_higher_ord} depict the improvement from capturing the higher-order interactions. LC-VAE clearly generates the teeth, indicating that the mouth is open. On the contrary, the MLC-VAE is able to recover the unseen attribute combination.

\section{Model exploration}
\label{sec:ablation}
For the purpose of model exploration, we investigate the following extensions to our model:
\begin{itemize}
    \item Firstly, we probe the limits of the proposed framework in terms of seen attribute combinations. In other words, we investigate how many attribute combinations must be present in the training set, so that generation of all combination is still possible.
    \item Secondly, we benchmark our model on a dataset with 3 attributes.
    \item Lastly, we extend the model to handle unlabelled attributes.
\end{itemize}{}

\subsection{Multiple missing combinations}
In order to study how the performance of the proposed model scales with the number of missing attribute combinations, we ablate the model on the MNIST dataset. In particular, we create 5 different training sets, that are missing 3, 6, 9, 15 and 20 attribute combinations respectively. We train 5 models and report the average classification accuracy of the generated unseen combinations. The result on Table \ref{tab:results_mulitple_missing} indicate that the model performs reasonably well even after removing 9 attribute combinations from the training set.

\begin{table}[hbt!]
\centering    

\begin{tabular}{cccccc}

\hline
\textbf{Accuracy}&\multicolumn{5}{c}{\textbf{Number of missing combinations}}\\

&3/30 & 6/30 & 9/30  &15/30&20/30 \\
\hline
Digit &88.1&94.5&90&50.4&15.8\\
Color &93.3&82.7&88.4&87.2&81.2\\

\hline

\end{tabular}

\centering    
\caption{Effect of number of unseen combinations to the accuracy (\%). The number of unseen combinations is explored (w.r.t. the total number of $30$ combinations). The accuracy of the attribute classifier on unseen attribute combinations is reported. Notice that even when $9$ combinations are unseen, there is no significant loss in the attribute classification. }

\label{tab:results_mulitple_missing}
\end{table}

\subsection{Multiple attributes}
The presented model can extended to handle an arbitrary number of attributes. The general form is as follows:
\begin{align}
M(\bm{y}_1, \bm{y}_2, \dots, \bm{y}_N) =\nonumber\\
\sum_{i=1}^{N} \bm{W}^{[i]} \bm{y}_i+ 
\sum_{j=2}^{N} \sum_{i=1}^{j-1} \bm{\mathcal{W}}^{[ij]} \times_2 \bm{y}_i \times_3 \bm{y}_j +  \nonumber\\
\sum_{h=3}^{N} \sum_{j=2}^{h-1} \sum_{i=1}^{j-1} \bm{\mathcal{W}}^{[ijh]} \times_2 \bm{y}_i \times_3 \bm{y}_j \times_4 \bm{y}_h + \dots \nonumber \\
+ \bm{\mathcal{W}}^{[1\dots N]} \times_2 \bm{y}_1 \times_3 \bm{y}_2 \dots \times_N \bm{y}_N
\label{E:polynomial}
\end{align}

In order to explore generation of more than two sets of attributes, we utilize the Morpho-MNIST \cite{castro2019morphomnist} dataset. This dataset is an extension of MNIST that is further transformed and annotated with regards to stroke thickness, swelling and fractures. Our 3-attribute benchmark consists of the thick and thin subsets of Morpho-MNIST, colored according to our previous benchmarks.

\begin{figure}[] 
    \centering
    \includegraphics[width=0.9\linewidth]{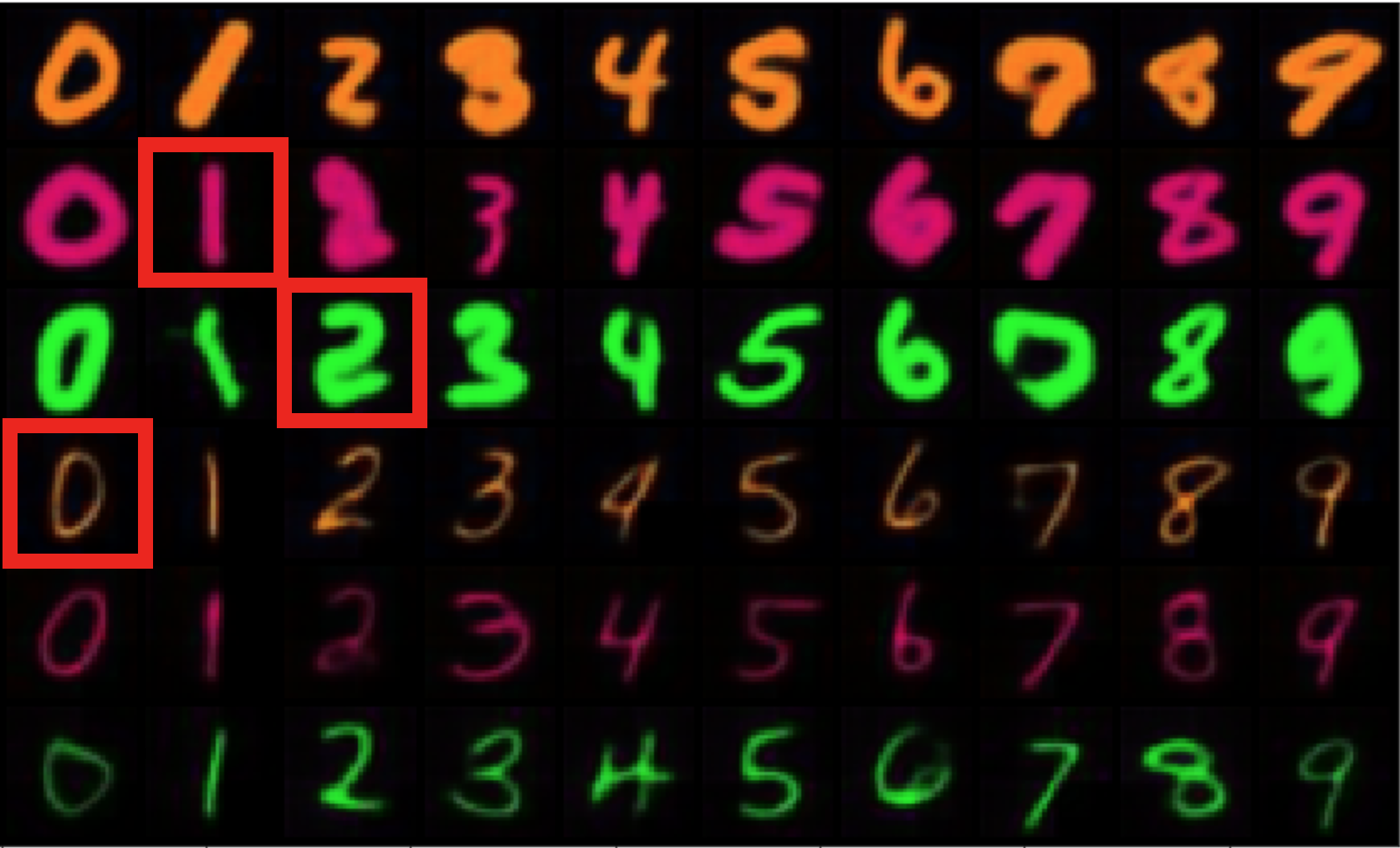}
    \caption{Samples synthesized on 3-attribute MNIST. The samples in the red boxes are combinations not seen during training (there are 3 combinations not seen in total). Nevertheless, the proposed method is able to synthesize images of all three unseen combinations.}
    \label{fig:3_atts}
\end{figure}

Based on \ref{E:polynomial}, the model for 3 attributes is:
\begin{align}
M(\bm{y}_1, \bm{y}_2,\bm{y}_3) =\nonumber\\
\bm{W}^{[1]} \bm{y}_1+ \bm{W}^{[2]} \bm{y}_2+ \bm{W}^{[3]} \bm{y}_3+\nonumber\\
\bm{\mathcal{W}}^{[12]} \times_2 \bm{y}_1 \times_3 \bm{y}_2 +
\bm{\mathcal{W}}^{[13]} \times_2 \bm{y}_1 \times_3 \bm{y}_3 +\nonumber\\
\bm{\mathcal{W}}^{[23]} \times_2 \bm{y}_2 \times_3 \bm{y}_3 +
\bm{\mathcal{W}}^{[123]} \times_2 \bm{y}_1 \times_3 \bm{y}_2 \times_4 \bm{y}_3,
\label{E:polynomial_3_atts}
\end{align}
where third order tensors $\bm{\mathcal{W}}^{[12]}, \bm{\mathcal{W}}^{[23]}, \bm{\mathcal{W}}^{[13]}$ capture all possible pairwise interactions, while $\bm{\mathcal{W}}^{[123]}$ is a fourth order tensor that captures the triplet-wise interaction between all attributes.
The synthetic results in Figure \ref{fig:3_atts} indicate that our model can effectively handle multiple attributes. 

\subsection{Unlabeled attributes}
In this section, we extend the model to handle unlabelled attributes. In particular, we assume prior knowledge of only the number of classes of the unlabelled attribute (e.g., 2 different colors). In this setting, the data have labels corresponding to one attribute but not the other. A second encoder network is utilized to infer the unobserved attribute's class probabilities. We utilize the Gumbel-Softmax distribution \cite{maddison2016concrete, jang2016categorical} to perform the reparametrization trick and train the model. 

\begin{figure}[] 
    \centering
    \includegraphics[width=0.9\linewidth]{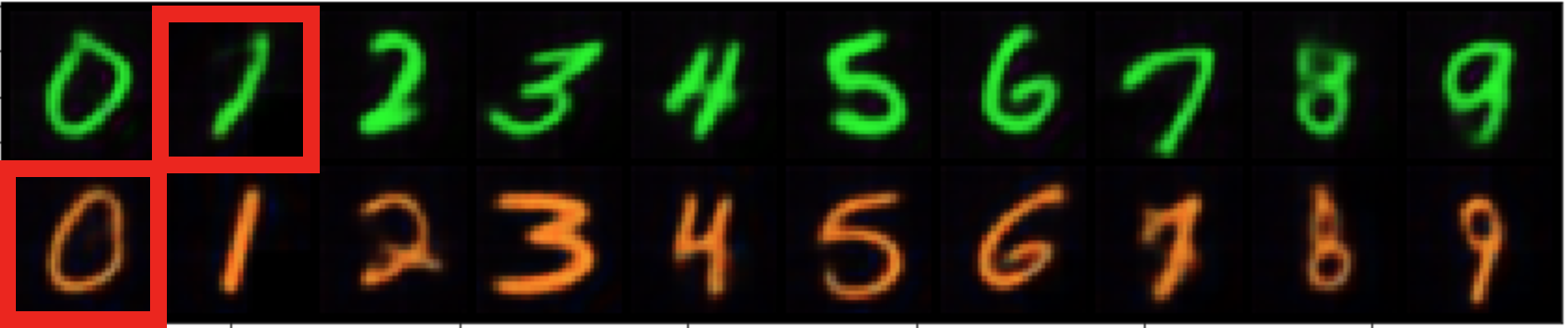}
    \caption{Results when color labels are not provided. Notice that the proposed method can still separate the attributes and synthesize the unseen combinations.}
    \label{fig:no_colour_label}
\end{figure}

We experiment with two-color MNIST (green/orange), without knowledge of the color label of each sample. Qualitative results are presented in Figure \ref{fig:no_colour_label}. Even though the images do not have color labels, the model is able to disentangle color from digit and generate the unseen attribute combinations (0-orange and 1-green).

\section{Conclusion}
In this work, we study the ability of conditional generative models to synthesize images of unseen attribute combinations. In particular, we focus on the setting where a number of attribute combinations are absent from the training set. Our goal is to learn a different conditional distribution for each class (i.e., attribute combination). The mean of each conditional distribution is learned through the proposed multilinear conditioning framework. We demonstrate the efficacy of our method on 3 different datasets and display the importance of modeling the interactions between labels. In the future, we want to explore further the multiplicative interactions of an arbitrarily large number of attributes, in the supervised and unsupervised setting, by using coupled decompositions and introducing weight sharing.

\bibliography{example_paper}
\bibliographystyle{icml2020}

\end{document}